\documentclass{article}

\usepackage{amsmath}
\usepackage{amsfonts}
\usepackage{amsthm}
\usepackage{amssymb}

\usepackage{fixmath}
\usepackage{mathtools}

\usepackage{subcaption}

\usepackage[numbers]{natbib}
\usepackage[final]{nips_2016}
\renewcommand{\vec}[1]{\mathbold{#1}}
\newcommand{\mat}[1]{\mathbold{#1}}
\newcommand{\matel}[1]{#1}
\newcommand{\tens}[1]{\boldsymbol{\mathcal{#1}}}
\newcommand{\tensel}[1]{\mathcal{#1}}

\usepackage[utf8]{inputenc} 
\usepackage[T1]{fontenc}    
\usepackage{hyperref}       
\usepackage{url}            
\usepackage{booktabs}       
\usepackage{amsfonts}       
\usepackage{nicefrac}       
\usepackage{microtype}      

 \title{Ultimate tensorization:\\
 compressing convolutional and FC layers alike}

\author{
Timur Garipov$^{1}$  ~~~~ Dmitry Podoprikhin$^{1,2}$ ~~~~ Alexander Novikov$^{3,4}$ ~~~~ Dmitry Vetrov$^{2,3}$\\
$^1$Moscow State University, Moscow, Russia \\
$^2$Yandex, Moscow, Russia  \\
$^3$National Research University Higher School of Economics, Moscow, Russia \\
$^4$Institute of Numerical Mathematics of the Russian Academy of Sciences, Moscow, Russia  \\
\texttt{timgaripov@gmail.com} ~~~~ \texttt{podoprikhin.dmitry@gmail.com}\\
\texttt{novikov@bayesgroup.ru} ~~~~ \texttt{vetrovd@yandex.ru}
}

\begin{document}

\maketitle

\begin{abstract}
  Convolutional neural networks excel in image recognition tasks, but this comes at the cost of high computational and memory complexity. To tackle this problem,~\cite{novikov2015tensorizing}~developed a tensor factorization framework to compress fully-connected layers. In this paper, we focus on compressing convolutional layers.
  We show that while the direct application of the tensor framework~\cite{novikov2015tensorizing} to the 4-dimensional kernel of convolution does compress the layer, we can do better. We reshape the convolutional kernel into a tensor of higher order and factorize it. We combine the proposed approach with the previous work to compress both convolutional and fully-connected layers of a network and achieve $80\times$ network compression rate with $1.1\%$ accuracy drop on the CIFAR-10 dataset.
\end{abstract}

\section{Introduction}
Convolutional Neural Networks (CNNs) show state-of-the-art performance on many problems in computer vision, natural language processing and other fields~\cite{krizhevsky2012imagenet, kalchbrenner2014convolutional}. At the same time, CNNs require millions of floating point operations to process an image and therefore real-time applications need powerful CPU or GPU devices. Moreover, these networks contain millions of trainable parameters and consume hundreds of megabytes of storage and memory bandwidth~\cite{simonyan15}. Thus, CNNs are forced to use RAM instead of solely relying on the processor cache -- orders of magnitude more energy efficient memory device~\cite{han2015deep} --  which increases the energy consumption even more. These reasons restrain the spread of CNNs on mobile devices.

To address the storage and memory requirements of neural networks,~\cite{novikov2015tensorizing}  used tensor decomposition techniques to compress fully-connected layers. They represented the parameters of the layers in the Tensor Train format~\cite{oseledets2011ttMain} and learned the network from scratch in this representation. This approach provided enough compression to move the storage bottleneck of VGG-16~\cite{han2015deep} from the fully-connected layers to convolutional layers. For a more detailed literature overview, see Sec.~\ref{sec:related-works}.

In this paper, we propose a tensor factorization based method to compress convolutional layers. Our contributions are:
\begin{itemize}
    \item We experimentally show that applying the Tensor Train decomposition -- the compression technique used in~\cite{novikov2015tensorizing} -- directly to the tensor of a convolution yields poor results (see Sec.~\ref{sec:experiments}). We explain this behavior and propose a way to reshape the 4-dimensional kernel of a convolution into a multidimensional tensor to fully utilize the compression power of the Tensor Train decomposition (see Sec.~\ref{sec:tt-conv}).
    \item We experimentally show that the proposed approach allows compressing a network that consists only of convolutions up to $4\times$ times with $2\%$ accuracy decrease (Sec.~\ref{sec:experiments}).
    \item We combine the proposed approach with the fully-connected layers compression of~\cite{novikov2015tensorizing}. Compressing both convolutional and fully-connected layers of a network yields $82\times$ network compression with $1\%$ accuracy drop, see Sec.~\ref{sec:experiments}.
\end{itemize}

\section{Convolutional Layer}
\label{sec:conv}
A convolutional network is a type of feed-forward architecture that transforms an input image to the final class scores using a sequence of layers. The main building block of such networks is a convolutional layer, that transforms the $3$-dimensional input tensor $\tens{X} \in \mathbb{R}^{W \times H \times C}$ into the output tensor $\tens{Y} \in \mathbb{R}^{W-l + 1\times H-l  + 1\times S}$ by \emph{convolving} $\tens{X}$ with the kernel tensor $\tens{K} \in \mathbb{R}^{\ell \times \ell \times C \times S}$:
\begin{equation}
\label{eq:conv}
    \tensel{Y}(x,y,s) = \sum\limits_{i= 1}^{\ell} \sum\limits_{j= 1}^{\ell}\sum\limits_{c = 1}^{C} \tensel{K}(i, j, c, s)\, \tensel{X}(x + i - 1, y + j - 1, c).
\end{equation}

\begin{figure}[ht]
    \caption{Reducing convolution~\eqref{eq:conv} to a matrix-by-matrix multiplication. \label{fig:conv-to-mat}}
    \centering
    \includegraphics[width=0.7\textwidth]{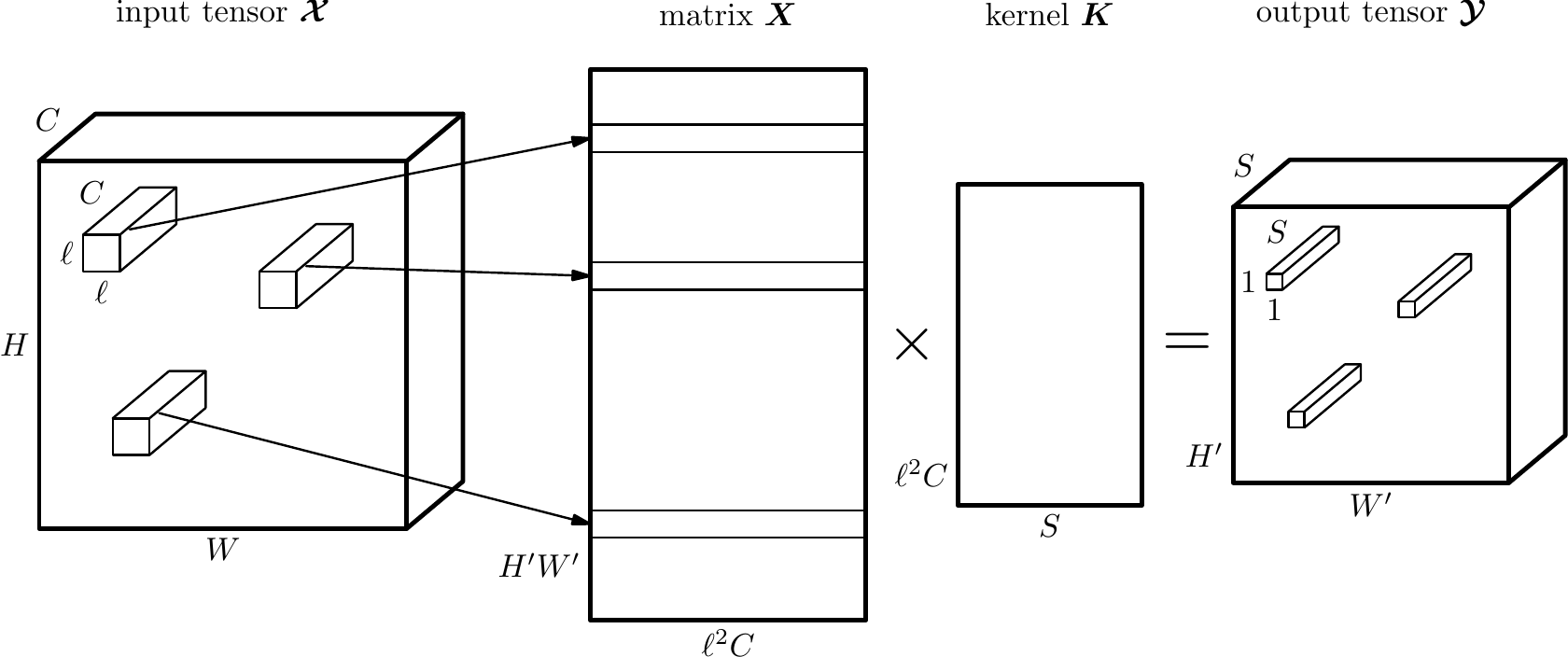}
\end{figure}

 To improve the computational performance, many deep learning frameworks reduce the convolution~\eqref{eq:conv} to a matrix-by-matrix multiplication~\cite{jia2014caffe, vedaldi2015matconvnet} (see Fig.~\ref{fig:conv-to-mat}). We exploit this matrix formulation to motivate a particular way of applying the Tensor Train format to the convolutional kernel (see Sec.~\ref{sec:tt-conv}). In the rest of this section, we introduce the notation needed to reformulate convolution~\eqref{eq:conv} as a matrix-by-matrix multiplication~$\mat{Y}=\mat{X}\mat{K}$.

 For convenience, we denote $H' = H- \ell + 1$ and $W' = W - \ell + 1$. Let us reshape the  output tensor $\tens{Y} \in \mathbb{R}^{W' \times H' \times S}$ into a matrix $\mat{Y}$ of size $W' H' \times S $ in the following way
$$\tensel{Y}(x,y,s) = Y(x + W'(y-1),s).$$

Let us introduce a matrix $\mat{X}$ of size $W'H'\times \ell^2C$, the $k$-th row of which corresponds to the $\ell \times \ell \times C$ patch of the input tensor that is used to compute the $k$-th row of the matrix $\mat{Y}$
$$ \tensel{X}(x + i - 1, y + j - 1, c) = X(x + W'(y-1), i + \ell(j-1) + \ell^2(c-1)),$$
where $y = 1,\ldots,H'$,\, $x = 1, \ldots, W'$,\, $i,j = 1,\ldots,\ell$.
Finally, we reshape the kernel tensor $\tens{K}$ into a matrix $\mat{K}$ of size $\ell^2C \times S$
$$\tensel{K}(i,j,c,s) =  K(i + \ell(j-1) + \ell^2(c-1), s).$$
Using the matrices defined above, we can rewrite the convolution definition~\eqref{eq:conv} as $\mat{Y}=\mat{X}\mat{K}$.

Note that the compression approach presented in the rest of the paper works with other types of convolutions, such as convolutions with padding, stride larger than $1$, or rectangular filters. But for clarity, we illustrate the proposed idea on the basic convolution~\eqref{eq:conv}.

\section{Tensor Train Decomposition}
\label{sec:TT-format}
The TT-decomposition (or TT-representation) of a tensor $\tens{A} \in \mathbb{R}^{n_1 \times \ldots \times n_d}$ is the set of matrices $\mat{G}_k[j_k] \in \mathbb{R}^{r_{k-1}\times r_{k}},$ where $j_k=1,\ldots,n_k, \, k=1,\ldots,d,$ and $r_0 = r_d = 1$, such that each of the tensor elements can be represented as
\begin{equation}
\label{eq:TT-format}
\tensel{A}(j_1,j_2,\ldots,j_d) = \mat{G}_1[j_1]\mat{G}_2[j_2]\ldots\mat{G}_d[j_d].
\end{equation}
The elements of the collection $\{r_k\} _{k=0}^{d}$  are called \emph{TT-ranks}. The collections of matrices $\{\{\mat{G}_k[j_k ]\}_{j_k = 1}^{n_k}\}_{k = 1}^d$ are called \emph{TT-cores} \cite{oseledets2011ttMain}.

The TT-format requires $\sum_{k=1}^d n_k r_{k-1}r_k$ parameters to represent a tensor $\tens{A} \in \mathbb{R}^{n_1 \times \ldots \times n_d}$ which has $\prod_{k=1}^d n_k$ elements.
The TT-ranks $r_k$ control the trade-off between the number of parameters versus the accuracy of the representation: the smaller the TT-ranks, the more memory efficient the TT-format is. An attractive property of the TT-format is the ability to efficiently perform basic linear algebra operations on tensors by working on the TT-cores of the TT-format, i.e. without materializing the tensor itself~\cite{oseledets2011ttMain}.

For a matrix -- a $2$-dimensional tensor -- the TT-decomposition  coincides with the matrix low-rank decomposition. To represent a matrix more compactly than in the low-rank format, the matrix TT-format is defined in a special way. Let us consider a matrix $\mat{A}$ of size $M \times N,$ where $M~=~\prod_{k=1}^{d}m_k, N =  \prod_{k=1}^{d}n_k$,
and reshape it into a tensor $\tens{A}$ of size $n_1 m_1 \times n_2 m_2 \times \ldots \times n_d m_d$ by defining  bijective mappings $\vec{\mu}(\ell) = (\mu_1(\ell),\ldots,\mu_d(\ell))$ and $\vec{\nu}(t) = (\nu_1(t),\ldots,\nu_d(t))$. The mapping
$\vec{\mu}(\cdot)$ maps row index $\ell = 1,\ldots,M$ into a $d$-dimensional vector index, where $k$-th dimension $\mu_k(\cdot)$ varies from $1$ to $m_k$.
The bijection  $\vec{\nu}(\cdot)$ maps column index $t=1,\ldots,N$ into a $d$-dimensional vector index, where $k$-th dimension $\nu_k(\cdot)$ varies from $1$ to $n_k$. Thus, using these mappings, we can form the tensor $\tens{A}$, whose $k$-th dimension is indexed by the compound index $(\mu_k(\cdot), \nu_k(\cdot)),$ and consider its TT-representation:
$$ A(\ell,t) = \tensel{A}((\mu_1(\ell),\nu_1(t)),\ldots,(\mu_d(\ell), \nu_d(t))) = \mat{G}_1[(\mu_1(\ell),\nu_1(t))]\ldots\mat{G}_d[(\mu_d(\ell),\nu_d(t))].$$

\section{TT-convolutional Layer \label{sec:tt-conv}}
In this section, we propose two ways to represent a convolutional kernel $\tens{K}$ in the TT-format.
One way is to apply the TT-decomposition to the tensor $\tens{K}$ directly. To see the drawbacks of this approach, consider a $1\times1$ convolution, which is a small fully-connected layer applied to the channels of the input image in each pixel location. The kernel of such convolution is essentially a $2$-dimensional array, and the TT-decomposition of $2$-dimensional arrays coincides with the matrix low-rank format. But for fully-connected layers, the matrix TT-format proved to be more efficient than the matrix low-rank format \cite{novikov2015tensorizing}. Thus, we seek for a decomposition that would coincide with the matrix TT-format on $1\times1$ convolutions.

Taking into account that a convolutional layer can be formulated as a matrix-by-matrix multiplication (see Sec.~\ref{sec:conv}), we reshape the 4-dimensional kernel tensor into a matrix  $\mat{K}$ of size  $\ell^2 C \times S$, where $\tensel{K}(x,y,c,s) =
K(\ell(y-1) +x + \ell^2(c-1),s)$. Then we apply the matrix TT-format (see Sec.~\ref{sec:TT-format}) to the matrix $\mat{K}$, i.e. reshape it into a tensor $\widetilde{\tens{K}}$ and convert it into the TT-format.
To reshape the matrix $\mat{K}$ into a tensor, we assume that its dimensions factorize: $C = \prod_{i = 1}^d C_i$ and $S = \prod_{i = 1}^d S_i$.
This assumption us not restrictive since we can always add some dummy channels filled with zeros to increase the values of $C$ and $S$.
Then we can define a $(d+1)$-dimensional tensor, where $k$-th dimension has the length $C_k S_k$ for $k = 1, \ldots, d$ and $\ell^2$ for $k=0$. Thus we obtain the following representation of the matrix $\mat{K}$
\begin{multline}
\label{eq:conv1}
\matel{K}(x  + \ell(y-1) + \ell^2(c'-1), s')  =
\widetilde{\tensel{K}}(\,(x + \ell(y -1),1),\,(c_1,s_1),\ldots,\,(c_d,s_d)) = \\
= \widetilde{\mat{G}}_0[x + \ell(y-1),1]\mat{G}_1[c_1,s_1] \ldots \mat{G}_d[c_d,s_d],
\end{multline}
where $c' = c_1 + \sum_{i=2}^{d}(c_i -
1)\prod_{j=1}^{i-1}C_j$ and $s' = s_1 + \sum_{i=2}^{d}(s_i -
1)\prod_{j=1}^{i-1}S_j$.

To simplify the notation, we index the $0$-th core $\widetilde{\mat{G}}_0$ with $x$ and $y$: $\mat{G}_0[x,y] = \widetilde{\mat{G}}_0[\ell(y-1)+x,1],$ where $x,y = 1,\ldots,\ell.$ Finally, substituting $\widetilde{\mat{G}}_0$ into~\eqref{eq:conv1}, we obtain the following decomposition of the convolution kernel $\tens{K}$
\begin{equation}
\label{eq:tt-conv}
\tensel{K}(x,y,c', s')  = \mat{G}_0[x,y]\mat{G}_1[c_1,s_1]...\mat{G}_d[c_d,s_d].
\end{equation}

To summarize our pipeline starting from an input tensor $\tens{X}$ (an image): the TT-convolutional layer firstly reshapes the input tensor into a $(2+d)$-dimensional tensor  $\widetilde{\tens{X}}$ of size $W\times H \times C_1 \times \ldots \times C_d$; then, the layer transforms the input tensor $\widetilde{\tens{X}}$ into the output tensor $\widetilde{\tens{Y}}$ of size $(W-\ell + 1) \times (H - \ell + 1) \times S_1 \ldots \times S_d$ in the following way
\begin{align*}
\widetilde{\tensel{Y}}(x,y,s_1,\ldots,s_d) = \sum\limits_{i=1}^{\ell}\sum\limits_{j=1}^{\ell} \sum\limits_{c_1,...,c_d} \,\,&\widetilde{\tens{X}}(i+x - 1, j+y - 1,c_1,\ldots,c_d)\\
&\mat{G}_0[i,j] \mat{G}_1[c_1,s_1]\ldots \mat{G}_d[c_d,s_d].
\end{align*}

Note that for a $1 \times 1$ convolution ($\ell = 1$), the $x$ and $y$ indices vanish from the decomposition~\eqref{eq:tt-conv}. The convolutional kernel collapses into a matrix~$\{\tensel{K}(1, 1, c', s')\}_{c' = 1, s'=1}^{C, S}$ and the decomposition~\eqref{eq:tt-conv} for this matrix coincides with the Tensor Train format for the fully-connected layer proposed in~\cite{novikov2015tensorizing}.

To train a network with TT-conv layers, we treat the elements of the TT-cores as the parameters of the layer and apply stochastic gradient descent with momentum to them. To compute the necessary gradients we use automatic differentiation implemented in TensorFlow~\cite{tensorflow2015-whitepaper}.

\section{Related Work \label{sec:related-works}}
Fully-connected layers of neural networks are traditionally considered as the memory bottleneck and numerous works focused on compressing these layers~\cite{chen2015compressing, xue2013restructuring, novikov2015tensorizing, sainath2013low}.
However, several state-of-the-art neural networks are either bottlenecked by convolutional layers~\cite{szegedy2015going, he2015deep}, or their fully-connected layers can be compressed to move the bottleneck to the convolutional layers~\cite{novikov2015tensorizing}. This leads to a number of works focusing on compressing and speeding up the convolutional layers~\cite{han2015deep, han2015learning, Denil2013predicting, figurnov2016perforatedcnns, lebedev2015speeding, wang2016factorized}.

One approach to compressing a convolutional layer is based on either pruning less important weights from the convolutional kernel, or restricting possible variation of the weights (quantization), or both~\cite{han2015deep, han2015learning, figurnov2016perforatedcnns}. Our approach is compatible with the quantization technique: one can quantize the elements of the TT cores of the decomposition. Some works also add Huffman coding on top of other compression techniques~\cite{han2015deep}, which is also compatible with the proposed method.

Another approach is to use tensor or matrix decompositions. CP-decomposition~\cite{lebedev2015speeding} and Kronecker product factorization~\cite{wang2016factorized} allow to speed up the inference time of convolutions and compress the network as a side effect.

\section{Experiments \label{sec:experiments}}
We evaluated the compressing strength of the proposed approach on CIFAR-10 dataset~\cite{krizhevsky2009learning}, which has $50\,000$ train images and $10\,000$ test images.
In all the experiments, we used stochastic gradient descent with momentum with coefficient~$0.9$, trained for~$100$ epochs starting from the learning rate of~$0.1$ and decreased it~$10\times$ after each~$30$ epochs.
To make the experiments reproducible, we released the codebase\footnote{\url{https://github.com/timgaripov/TensorNet-TF}}.

We used two architectures as references: the first one is dominated by the convolutions (they occupy $99.54\%$ parameters of the network), and the second one is dominated by the fully-connected layers (they occupy $95.98\%$ parameters of the network).




\paragraph{Convolutional network.}
The first network has the following architecture:
conv ($64$ output channels); BN; ReLU; conv ($64$ output channels); BN; ReLU; max-pool ($3 \times 3$ with stride $2$); conv ($128$ output channels); BN; ReLU; conv ($128$ output channels); BN; ReLU; max-pool ($3 \times 3$ with stride $2$); conv ($128$ output channels); BN; ReLU; conv ($128$ output channels); avg-pool ($4 \times 4$); fc ($128 \times 10$), where 'BN' stands for batch normalization \cite{ioffe2015batch} and all convolutional filters are of size $3\times 3$.
To compress the network we replace each convolutional layer excluding the first one (it contains less than $1\%$ of the network parameters) with the TT-conv layer~(see Sec.~\ref{sec:tt-conv}). For training, we initialize the TT-cores of the TT-conv layers with random noise and train the whole network from scratch.

We compare the proposed TT-convolution against the naive approach -- directly applying the TT-decomposition to the $4$-dimensional convolutional kernel (see Sec.\ref{sec:tt-conv}). We report that on the $2\times$ compression level the proposed approach ($0.8\%$ loss of accuracy) outperforms the naive baseline ($2.4\%$ loss of accuracy), for details see Tbl.~\ref{tbl:compression}a.

\begin{table}
\centering
\caption{Compressing convolutional networks on the CIFAR-10 dataset (see Sec.~\ref{sec:experiments}). Different rows with the same model name correspond to different choices of the TT-ranks\label{tbl:compression}.}
\begin{subtable}{.45\textwidth}
\centering

\caption{Compressing the first baseline (`conv'), which is dominated by convolutions. `TT-conv': the proposed compression method; `TT-conv (naive)': direct application of the TT-decomposition to convolutional kernels.}

\begin{tabular}{lll}
    \toprule
    Model    & top-1 acc. & compr.  \\
    \midrule
    conv (baseline) & $90.7$ & $1$ \\
    TT-conv & $89.9$  & $2.02$   \\
    TT-conv & $89.2$ & $2.53$  \\
    TT-conv & $89.3$ & $3.23$   \\
    TT-conv & $88.7$ & $4.02$  \\
    TT-conv (naive) & $88.3$ & $2.02$ \\
    TT-conv (naive) & $87.6$ & $2.90$ \\
    \bottomrule
  \end{tabular}
\end{subtable}
\hfill
\begin{subtable}{.45\textwidth}\centering
  \caption{Compressing the second baseline (`conv-fc'), which is dominated by fully-connected layers. `conv-TT-fc': only the fully-connected part of the network is compressed; `TT-conv-TT-fc': fully-connected and convolutional parts are compressed.}

  \begin{tabular}{llll}
    \toprule
    Model    & top-1 acc. & compr.\\
    \midrule
    conv-fc (baseline) & $90.5$ & $1$  \\
    conv-TT-fc & $90.3$ & $10.72$   \\
    conv-TT-fc & $89.8$ & $19.38$  \\
    conv-TT-fc & $89.8$ & $21.01$ \\
    TT-conv-TT-fc & $90.1$ & $9.69$  \\
    TT-conv-TT-fc & $89.7$ & $41.65$  \\
    TT-conv-TT-fc & $89.4$ & $82.87$  \\
    \bottomrule
  \end{tabular}
 \end{subtable}
\end{table}

\paragraph{Network with convolutions and fully-connected layers.}
The second reference network was obtained from the first one by replacing the average pooling with two fully-connected layers of size $8192 \times 1536$ and $1536 \times 512$.

To compress the second network, we replace all layers excluding the first and the last one (they occupy less than $1\%$ of parameters) with TT-conv and TT-fc~\cite{novikov2015tensorizing} layers.
To speed up the convergence, we trained the network in two stages:
first we replaced only the convolutional layers with TT-conv layers and trained the network; then we replaced the fully-connected layers with randomly initialized TT-fc layers and fine-tuned the whole model.

To compare against~\cite{novikov2015tensorizing} we include the results of compressing only fully-connected layers (Tbl.~\ref{tbl:compression}b).
Initially, the fully-connected part was the memory bottleneck and it was more fruitful to compress it while leaving the convolutions untouched: we obtained $10.72\times$ network compression with $0.2\%$ accuracy drop by compressing only fully-connected layers, and $9.61\times$ compression with $0.4\%$ accuracy drop by compressing both fully-connected and convolutional layers.
But after the first gains, the bottleneck moved to the convolutional part and the fully-connected layers compression capped at about $21\times$ network compression.
At this point, by additionally factorizing the convolutions we raised the network compression up to $80\times$ while losing $1.1\%$ of accuracy~(Tbl.~\ref{tbl:compression}b).

\section{Conclusion}
In this paper, we proposed a tensor decomposition approach to compressing convolutional layers of a neural network. By combing this convolutional approach with the work~\cite{novikov2015tensorizing} for fully-connected layers, we compressed a convolutional network $80\times$ times. These results make a step towards the era of embedding compressed models into smartphones to allow them constantly look and listen to its surroundings.

In the future work, we will experiment with the proposed approach on the ILSVRC-2012 dataset~\cite{Russakovsky2015ImageNet} on state-of-the-art neural architectures.

\small{
\bibliography{tensor-net}
\bibliographystyle{unsrt}

}

\end{document}